\documentclass[conference]{IEEEtran}
\IEEEoverridecommandlockouts
\usepackage{cite}
\usepackage{amsmath,amssymb,amsfonts}
\usepackage{graphicx}
\usepackage{textcomp}
\usepackage{xcolor}
\usepackage{pifont}
\usepackage{bbding}
\usepackage{authblk}
\usepackage{xspace}
\usepackage{booktabs}
\usepackage{multirow}
\usepackage{algorithm,algpseudocode}
\usepackage{placeins}
\usepackage{hyperref}
\hypersetup{hidelinks}
\def\BibTeX{{\rm B\kern-.05em{\sc i\kern-.025em b}\kern-.08em
    T\kern-.1667em\lower.7ex\hbox{E}\kern-.125emX}}

\makeatletter
\DeclareRobustCommand\onedot{\futurelet\@let@token\@onedot}
\def\@onedot{\ifx\@let@token.\else.\null\fi\xspace}

\makeatother

\begin{document}

\title{SMAFormer: Synergistic Multi-Attention Transformer for Medical Image Segmentation
}




\author[12]{Fuchen Zheng}
\author[123]{Xuhang Chen}
\author[1]{Weihuang Liu}
\author[1]{Haolun Li}
\author[1]{\authorcr Yingtie Lei}
\author[24]{Jiahui He}
\author[1*]{Chi-Man Pun\thanks{* Corresponding Authors.}}
\author[2*]{Shoujun Zhou}
\affil[1]{University of Macau}
\affil[2]{Shenzhen Institutes of Advanced Technology, Chinese Academy of Sciences}
\affil[3]{Huizhou University}
\affil[4]{University of Nottingham Ningbo China}

\maketitle

\begin{abstract}
Medical image segmentation has benefited from attention-based Transformers and residual U-shaped networks, yet small organs and irregular tumors remain difficult because local boundary cues are easily weakened by deep abstraction and repeated feature reshaping. We introduce SMAFormer, a Transformer-based segmentation architecture that coordinates complementary attention views while preserving fine spatial evidence.
SMAFormer contains two core components. The Synergistic Multi-Attention (SMA) Transformer block combines pixel, channel, and spatial attention to couple local responses with global context. The Feature Fusion Modulator reduces information loss during attention transitions and encoder-decoder feature fusion, strengthening the interaction between channel-level and spatial evidence.
Across multi-organ, liver tumor, and bladder tumor segmentation benchmarks, SMAFormer achieves strong reported results, including 96.07\% average DSC on ISICDM2019, 94.11\% on LiTS2017, and 86.08\% on Synapse. Code and models are available at: \url{https://github.com/lzeeorno/SMAFormer}.
\end{abstract}


\section{Introduction}

Early tumor diagnosis is important because late-stage cancer is often difficult to treat~\cite{1}. Artificial intelligence, and medical image segmentation in particular, can support this process by delineating lesions and anatomical structures from clinical images~\cite{chen11,chen9,chen6,chen5}. However, tiny tumors and small organs are often represented by weak boundary evidence. This evidence can be diluted when deep networks repeatedly downsample, reshape, and fuse feature maps. The resulting challenge is therefore twofold: preserving fine local structure while maintaining the global anatomical context needed for stable segmentation.

Recent convolutional neural network (CNN) based methods~\cite{chen12,chen13,chen14,chen10,py1,py2,py3,zhang4} have shown promise, but their repeated local operations may still suppress small-object features. Multi-attention mechanisms~\cite{chen1,chen2,chen3,chen4,chen7,chen8,zhang9,li1,li2} partly address this limitation. In many designs, however, different attention branches are added without an explicit mechanism for coordinating spatial, channel, and pixel-level evidence across scales.

This paper introduces \textbf{SMAFormer}, a Transformer-based model for medical image segmentation. Inspired by ResUNet~\cite{6resunet}, SMAFormer combines Transformer blocks with a U-shaped residual structure to retain multi-resolution information while coordinating complementary attention responses.

Our key contributions are:

\begin{enumerate}
\item \textbf{SMAFormer Architecture:} A residual U-shaped Transformer model that combines pixel, channel, spatial, and multi-head self-attention for medical image segmentation.
\item \textbf{Learnable Segmentation Modulator:} An embeddable module that supports multi-scale feature fusion and reduces information loss during attention transitions.
\item \textbf{Benchmark Evaluation:} Experiments on multi-organ, liver tumor, and bladder tumor segmentation datasets show consistent improvements over the compared methods in the reported metrics.
\end{enumerate}

\section{Related Work}

\subsection{Medical Image Segmentation}
Medical image segmentation involves partitioning medical images into clinically meaningful regions. The U-Net architecture~\cite{4unet} is widely used because it combines low-level spatial detail with high-level contextual information. Several variants, including ResUNet~\cite{6resunet}, introduce residual connections from ResNet~\cite{7resnet} to improve gradient flow and feature reuse. SMAFormer follows this residual U-shaped principle, but uses it as the backbone for coordinating multiple attention responses.


\subsection{Vision Transformer}
Transformers use global self-attention to model long-range dependencies in images, as demonstrated by Vision Transformer (ViT)~\cite{34vit}. This design has been adapted for medical imaging, with models such as Swin Transformer~\cite{12swin2021} and hybrid ResNet-Transformer architectures such as ResT~\cite{resT2022v2}.

In medical image segmentation, attention mechanisms have been integrated into U-Net architectures, giving rise to U-shaped Transformers such as TransUNet~\cite{15chen2021transunet}, which combines CNN features with Transformer-based global context. A remaining challenge is how to preserve local evidence while merging attention responses produced at different scales. SMAFormer addresses this gap through a residual U-shaped Transformer structure, an SMA block, and a feature fusion modulator.

\section{Method}
\begin{figure*}[ht]
\centerline{\includegraphics[width=0.8\linewidth]{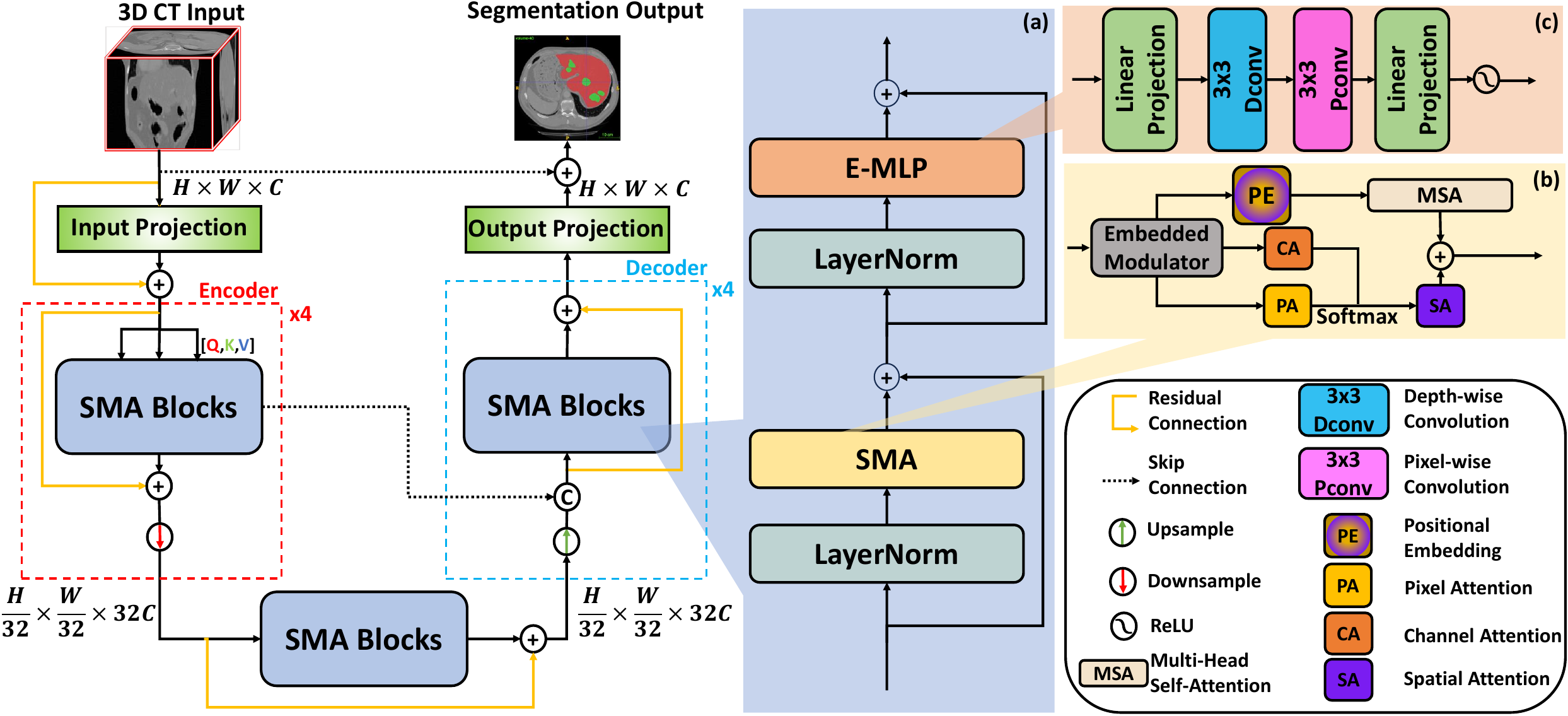}}
\caption{This figure provides an overview of the SMAFormer architecture. The figure details (a) the SMA Transformer block, (b) the SMA Part within the SMA Transformer block, and (c) the E-MLP Part within the SMA Transformer block.}
\label{fig1}
\end{figure*}


\subsection{Overview}
SMAFormer, as depicted in Figure~\ref{fig1}, adopts a hierarchical U-shaped architecture reminiscent of ResU-Net~\cite{53resunet++,66wang2022uformer}, incorporating skip-connections and residual connections between the encoder and decoder for efficient information propagation.

Given an input image or slice $I \in \mathbb{R}^{C_0 \times H \times W}$, SMAFormer first extracts low-level features through an initial projection layer comprising a $3 \times 3$ convolution followed by a ReLU activation. The extracted features are then passed through a four-stage encoder, mirroring the U-Net structure. Each encoder stage consists of a stack of SMA Transformer blocks (detailed in Section~\ref{sec:sma}) for capturing multi-scale features, followed by a down-sampling layer.

The down-sampling layer performs two operations. First, it records positional information within the embedded modulator (discussed in Section~\ref{sec:mod}). Second, it uses a residual convolution block consisting of three $3\times3$ convolutions with a stride of 2 to reduce the spatial dimensions while increasing the channel count. This residual down-sampling process preserves long-range dependencies. Specifically, given an input feature map $X_i \in \mathbb{R}^{C \times H \times W}$, the output of the i-th encoder stage is $X_{\text{conv}} + X_{\text{residual}} \in \mathbb{R}^{2^iC \times \frac{H}{2^i} \times \frac{W}{2^i}}$, where $X_{\text{conv}}$ denotes the convolved features and $X_{\text{residual}}$ represents the residual branch.

Mirroring the encoder, the decoder comprises four symmetrical stages. Each stage begins with a $2\times2$ transposed convolution to upsample the feature maps, effectively halving the channel count and doubling the spatial dimensions. Subsequently, the upsampled features are concatenated with the corresponding encoder features via skip connections, facilitating the fusion of high-level semantic information with low-level spatial details. Finally, an output convolution layer processes the concatenated features to generate the segmentation prediction.

\subsection{SMA Block}\label{sec:sma}
Directly applying conventional Transformers~\cite{zhang2} to medical image segmentation presents two challenges. First, attention may not concentrate on medically relevant regions when the target structures are small or low contrast. Second, global token mixing alone can underrepresent local boundary context, which is essential for segmenting small organs and irregular tumors.

To address these challenges, we introduce the Synergistic Multi-Attention (SMA) Transformer block, illustrated in Figure~\ref{fig1}. The block combines three attention mechanisms with multi-head self-attention so that local, channel-wise, and spatial evidence can be fused within the same representation.

\subsubsection{Synergistic Multi-Attention (SMA)}
Unlike approaches that restrict self-attention within local windows~\cite{66wang2022uformer}, SMA combines channel attention, spatial attention, and pixel attention with multi-head self-attention. As shown in Figure~\ref{fig1}(b), this design lets the model preserve fine spatial responses while still using global contextual cues. Given a feature map $X \in \mathbb{R}^{C \times H \times W}$, SMA divides it into patches and flattens the channels. The resulting features are processed by channel, pixel, and spatial attention branches in parallel. The pixel and channel attention outputs are combined through matrix multiplication and then refined by the spatial attention branch. The three branches are finally fused to generate the attention map.

\subsubsection{Enhanced Multi-Layer Perceptron (E-MLP)}
Standard Feed-Forward Networks (FFNs) have limited ability to model neighboring pixels~\cite{70MLP2021cvt,71eighboringpixels2021CVPR}. We therefore enhance the E-MLP by incorporating depth-wise and pixel-wise convolutions~\cite{72depth-wiseConv2021localvit,73depth-wiseConv2018mobilenetv2,74depth-wiseConv2021incorporating}. As shown in Figure~\ref{fig1}(c), E-MLP first projects the input tokens to a higher-dimensional space. The tokens are then reshaped into 2D feature maps and processed by a $3\times3$ pixel-wise convolution followed by a $3\times3$ depth-wise convolution. The features are reshaped back into tokens, projected to the original channel dimension, and passed through a GELU activation~\cite{75GELU2016gaussian}.

Mathematically, the computation within an SMA Transformer block can be expressed as:
\begin{equation}
\begin{aligned}
& X_{i+1}'= \operatorname{SMA}(\operatorname{LN}(X_{i})) + X_{i}, \\
& X_{i+1} = \operatorname{E-MLP}(\operatorname{LN}(X_{i+1}')) + X_{i+1}',
\end{aligned}
\end{equation}
where $X_i$ represents the input features to the i-th block, $X_{i+1}'$ and $X_{i+1}$ are the outputs of the $\operatorname{SMA}$ and $\operatorname{E-MLP}$ modules respectively, and $\operatorname{LN}$ denotes layer normalization.

Together, SMA and E-MLP allow each Transformer block to combine global context with local boundary evidence.

\subsection{Multi-Scale Segmentation Modulator}\label{sec:mod}
\begin{figure}[ht]
\centerline{\includegraphics[width=0.7\columnwidth]{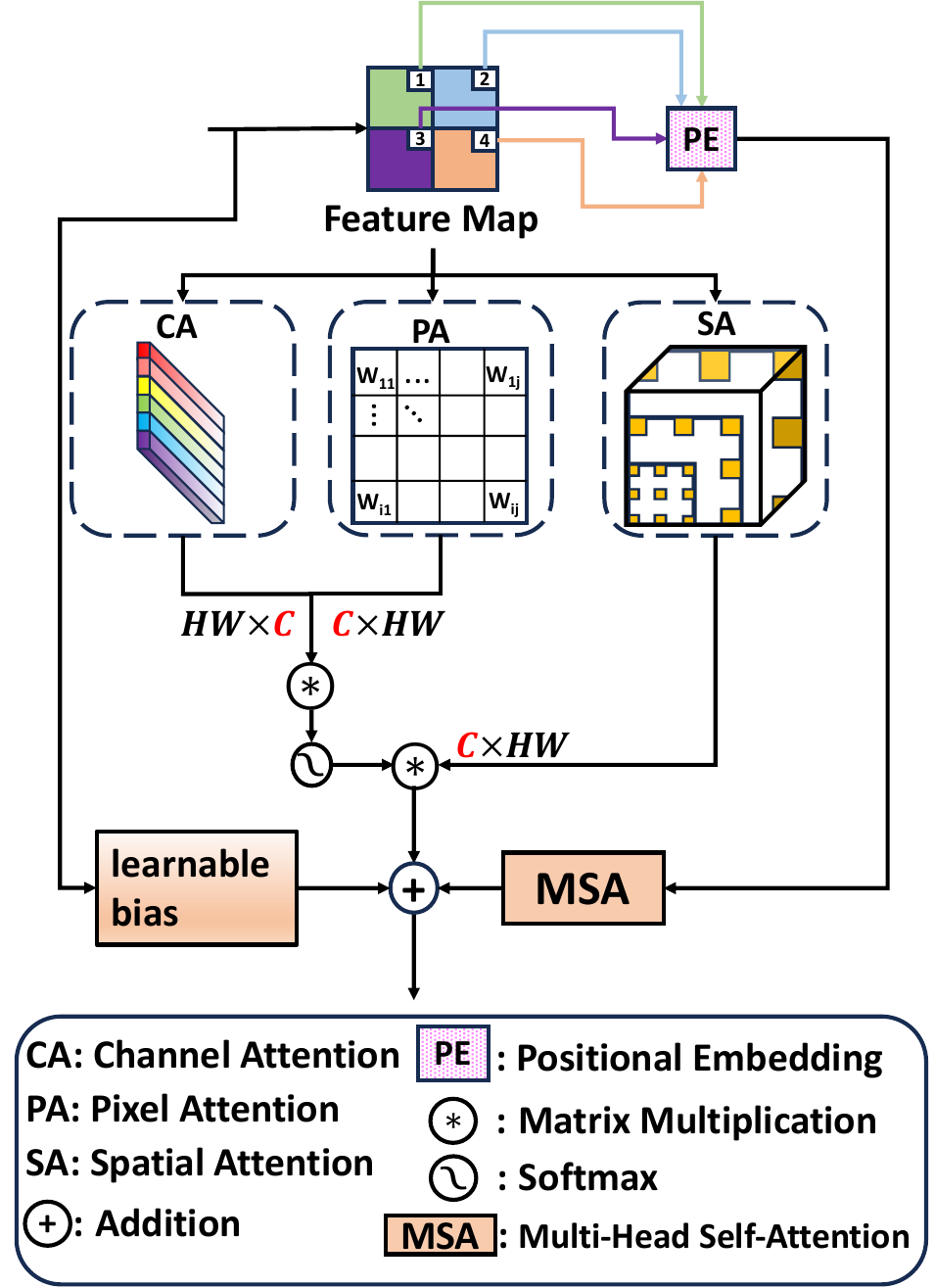}}
\caption{This figure presents a schematic diagram of the proposed modulator.}
\label{fig2}
\end{figure}

To further preserve fine-grained details during multi-scale attention fusion, we introduce the multi-scale segmentation modulator shown in Figure~\ref{fig2}. The figure summarizes the three functions used in the network: positional embedding for flattened feature maps, a trainable bias added to the attention output, and assistance for the transposition and matrix multiplication required by the channel, spatial, and pixel attention branches. In this way, the modulator links the multi-attention computation inside each SMA block with the encoder-decoder feature fusion process.

\subsection{Objective Function}
We train SMAFormer using the BCE Dice loss $\mathcal{L}_{BD}$~\cite{62BCEDiceLoss2016v}, a widely adopted loss function for segmentation tasks that combines the benefits of Binary Cross-Entropy (BCE) loss $\mathcal{L}_{BCE}$ and Dice loss $\mathcal{L}_{D}$:

\begin{equation}
\begin{aligned}
    \mathcal{L}_{BD} = & \mathcal{L}_{D} + \mathcal{L}_{BCE}(y, p) \\
    = & \frac{1}{N} \sum_{i=1}^{N} \left(1 - \frac{2 \sum_{j} y_{i,j} p_{i,j}}{\sum_{j} y_{i,j} + \sum_{j} p_{i,j}}\right) \\ & -{(y\log(p) + (1 - y)\log(1 - p))},
\end{aligned}
\end{equation}
where $y$ represents the ground truth segmentation mask, $p$ denotes the predicted segmentation mask, and $N$ is the number of pixels in the image. The BCE loss penalizes discrepancies between the predicted and true label distributions, while the Dice loss encourages overlap between the predicted and true segmentation regions. This combined loss function encourages both accurate pixel-wise classification and strong boundary delineation.

\section{Experiments}
\begin{table*}[ht]
\centering
\caption{Comparison with State-of-the-Art models on the ISICDM2019 and LiTS2017 datasets. The best results are bolded while the second best are underlined.}
\begin{tabular}{c|cccc|cccc}
\toprule
\multirow{3}{*}{Method}  & \multicolumn{4}{c|}{ISICDM2019}                                    & \multicolumn{4}{c}{LiTS2017}                                      \\ \cmidrule(l){2-9} 
                         & \multicolumn{2}{c}{Average}     & Bladder        & Tumor          & \multicolumn{2}{c}{Average}     & Liver        & Tumor          \\ \cmidrule(l){2-9} 
                         & DSC(\%) $\uparrow$        & mIoU(\%) $\uparrow$       & DSC(\%) $\uparrow$        & DSC(\%) $\uparrow$        & DSC(\%) $\uparrow$        & mIoU(\%) $\uparrow$       & DSC(\%) $\uparrow$        & DSC(\%) $\uparrow$        \\ 
                         \midrule
ViT~\cite{34vit}+CUP~\cite{15chen2021transunet}                  & 88.60          & 84.40          & 91.88          & 85.32          & 80.33          & 77.25          & 83.97          & 76.69          \\

R50-ViT~\cite{34vit}+CUP~\cite{15chen2021transunet}              & 88.77          & 85.62          & 92.05          & 85.49          & 82.62          & 79.68          & 85.83          & 79.41          \\

ResUNet++~\cite{53resunet++}                & 87.11          & 83.78          & 89.90          & 84.32          & 75.73 & 74.19 & 79.12 & 72.34             \\

ResT-V2-B~\cite{resT2022v2}                & 89.26          & 82.13          & 93.01          & 85.50          &     78.53 & 75.24 & 81.22 & 75.83           \\

TransUNet~\cite{15chen2021transunet}                & \underline{94.56} & \underline{93.60}  & \underline{97.74} & \underline{91.38}          &   \underline{93.28} & \underline{90.81}  & \underline{95.54} & \underline{91.03}               \\

SwinUNet~\cite{21swinUnet2022}                 & 91.95          & 89.77          & 94.73          & 89.17          &     89.68 & 86.62 & 93.31 & 86.04               \\
Swin UNETR~\cite{49swinUNETR}               & 92.60          & 90.61          & 95.08          & 90.12          &  91.95 & 90.02 & 94.73 & 89.17               \\
UNETR~\cite{48unetr}                    & 91.55          & 88.34          & 94.83          & 88.26          &  89.38 & 87.46 & 92.89 & 85.86               \\
nnFormer~\cite{47nnformer}                 & 93.54          & 89.11          & 96.97          & 90.41          &    91.74 & 89.95 & 94.57 & 88.91              \\ 
\midrule
\textbf{SMAFormer(Ours)} & \textbf{96.07} & \textbf{94.67} & \textbf{98.57} & \textbf{93.56} & \textbf{94.11} & \textbf{91.94} & \textbf{95.88} & \textbf{92.34} \\ 



\bottomrule
\end{tabular}
\label{tab:comp}
\end{table*}

Table~\ref{tab:comp} reports the primary tumor segmentation comparison on ISICDM2019 and LiTS2017. The table tests whether SMAFormer improves both average overlap and tumor-specific segmentation under the stated benchmark setting. On ISICDM2019, SMAFormer reaches 96.07\% average Dice and 94.67\% mIoU, compared with 94.56\% and 93.60\% for TransUNet. On LiTS2017, SMAFormer reaches 94.11\% average Dice and 91.94\% mIoU, compared with 93.28\% and 90.81\% for TransUNet. The tumor-class Dice values, 93.56\% on ISICDM2019 and 92.34\% on LiTS2017, directly address the small and irregular foreground regions that motivate the architecture.

\subsection{Datasets and Implementation Details}\label{sec:4a}

This work utilizes three publicly available medical image segmentation datasets: LiTS2017~\cite{44lits2017}, ISICDM2019~\cite{46ISICDM2019}, and Synapse~\cite{45synapse}. Following nnformer~\cite{47nnformer}, we employ an 80/15/5 train/validation/test split for all datasets. Input images are resized to $512\times512$ pixels.

SMAFormer is implemented in PyTorch and trained on a single NVIDIA GeForce RTX 4090 GPU using SGD with momentum (0.98), weight decay ($1e^{-6}$), and a cosine-decayed learning rate (initial: $1e^{-2}$, minimum: $6e^{-6}$). Data augmentation includes random horizontal flipping and rotation. We initialize the Transformer backbone with a ViT pre-trained model~\cite{77PretrainViT2020image}. 

\begin{table*}[ht]
    \centering
    \caption{Comparison with State-of-the-Art models on the Synapse multi-organ dataset. The best results are bolded while the second best are underlined.
    }
    \resizebox{\textwidth}{!}{%
    \begin{tabular}{c|c|c|c|c|c|c|c|c|c}
        \toprule
        Model  & \multicolumn{1}{c|}{Average} & Aorta & Gallbladder & Kidney(Left) & Kidney(Right) & Liver & Pancreas & Spleen & Stomach \\
               & DSC(\%)$\uparrow$ & DSC(\%)$\uparrow$ & DSC(\%)$\uparrow$ & DSC(\%)$\uparrow$ & DSC(\%)$\uparrow$ & DSC(\%)$\uparrow$ & DSC(\%)$\uparrow$ & DSC(\%)$\uparrow$ & DSC(\%)$\uparrow$\\
        \midrule

        ViT~\cite{34vit}+CUP~\cite{15chen2021transunet}    
            & 67.86 & 70.19 & 45.10 & 74.70 & 67.40 & 91.32 & 42.00 & 81.75 & 70.44     \\
    
        R50-ViT~\cite{34vit}+CUP~\cite{15chen2021transunet}  
            & 71.29 & 73.73 & 55.13 & 75.80 & 72.20 & 91.51 & 45.99 & 81.99 & 73.95     \\
            
        TransUNet~\cite{15chen2021transunet}
            & 84.36 & 90.68 & \underline{71.99} & \underline{86.04} & 83.71 & 95.54 & 73.96 & 88.80 & \textbf{84.20}      \\
            
        SwinUNet~\cite{21swinUnet2022}
            & 79.13 & 85.47 & 66.53 & 83.28 & 79.61 & 94.29 & 56.58 & \underline{90.66} & 76.60    \\

        UNETR~\cite{48unetr}
            & 79.56 & 89.99 & 60.56 & 85.66 & 84.80 & 94.46 & 59.25 & 87.81 & 73.99     \\



        
        Swin UNETR~\cite{49swinUNETR}
            & 73.51 & 82.94 & 60.96 & 80.41 & 71.14 & 91.55 & 56.71 & 77.46 & 66.94     \\
            
        CoTr~\cite{52cotr}
            & \underline{85.72} & \textbf{92.96} & 71.09 & 85.70 & 85.71 & \underline{96.88} & 81.28 & 90.44 & 81.74     \\

        nnFormer~\cite{47nnformer}
            & 85.32 & 90.72 & 71.67 & 85.60 & \underline{87.02} &  96.28 & \textbf{82.28} &  87.30 & 81.69     \\
        
        \midrule
        \textbf{SMAFormer(Ours)}
            & \textbf{86.08} & \underline{92.13} & \textbf{72.03} & \textbf{86.97} & \textbf{88.60} & \textbf{97.71} & \underline{81.93} & \textbf{91.77} & \underline{84.15}     \\

        \bottomrule
    \end{tabular}
    }
    \label{tab:syna}
\end{table*}

Table~\ref{tab:syna} reports Synapse multi-organ segmentation and tests whether the same architecture remains useful outside binary tumor segmentation. SMAFormer reaches 86.08\% average Dice, above TransUNet at 84.36\%, CoTr at 85.72\%, and nnFormer at 85.32\%. The organ-wise evidence is also important: SMAFormer reaches 72.03\% Dice on gallbladder, 81.93\% on pancreas, 97.71\% on liver, and 91.77\% on spleen. This pattern links the method back to the core motivation, because fine-grained attention should improve difficult structures without sacrificing global anatomical consistency.

\subsection{Evaluation Metrics} 
We evaluated the segmentation performance using two widely adopted metrics: \textbf{Dice Coefficient Score (DSC)} and \textbf{Mean Intersection over Union (mIoU)}.
\begin{equation}
    DSC = \frac{2 \times |P \cap G|}{|P| + |G|},
\end{equation}
where $P$ represents the predicted segmentation. $G$ represents the ground truth segmentation. $|P \cap G|$ represents the number of pixels in the intersection of the predicted and ground truth segmentations. $|P|$ represents the number of pixels in the predicted segmentation. $|G|$ represents the number of pixels in the ground truth segmentation.

\begin{equation}
    mIoU = \frac{1}{C} \sum_{i=1}^{C} \frac{|P_i \cap G_i|}{|P_i| + |G_i| - |P_i \cap G_i|},
\end{equation}
where $C$ represents the number of classes. $P_i$ represents the predicted segmentation for class $i$. $G_i$ represents the ground truth segmentation for class $i$.

\subsection{Comparisons with State-of-the-Art Methods}

This subsection compares SMAFormer with representative state-of-the-art segmentation methods (Tables~\ref{tab:comp} and~\ref{tab:syna}).

\begin{figure}[!t]
    \centering
    \includegraphics[width=0.98\columnwidth]{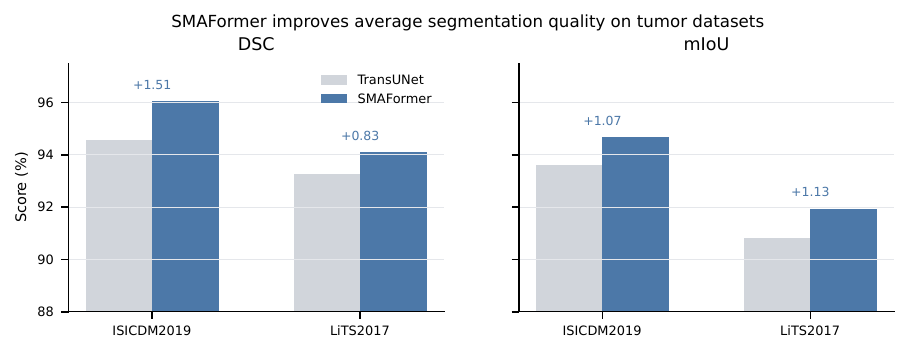}
    \caption{Bar chart of the average Dice and mIoU scores reported in Table~\ref{tab:comp}. SMAFormer consistently improves the strongest shared baseline across ISICDM2019 and LiTS2017.}   \label{fig:nature_smaformer_average}
\end{figure}

\begin{figure}[!t]
    \centering
    \includegraphics[width=0.98\columnwidth]{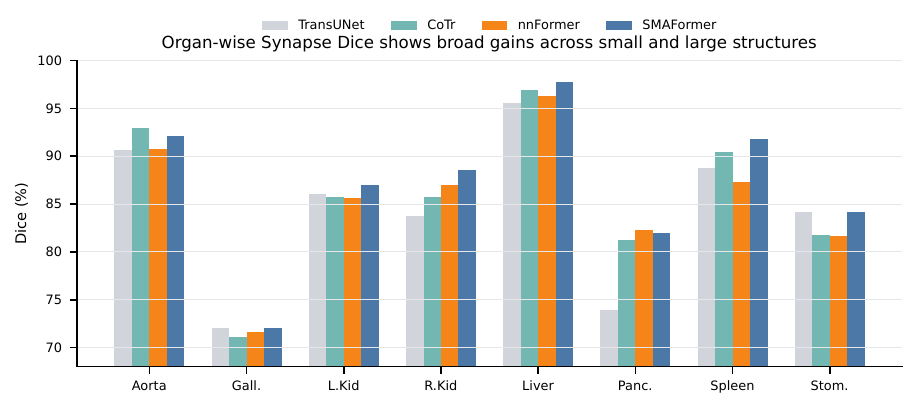}
    \caption{Organ-wise Synapse comparison from Table~\ref{tab:syna}. The visualization emphasizes where SMAFormer improves small and anatomically variable organs while retaining competitive performance on larger organs.}    \label{fig:nature_smaformer_synapse_organs}
\end{figure}

\begin{figure}[ht]
\centering
\includegraphics[width=\columnwidth]{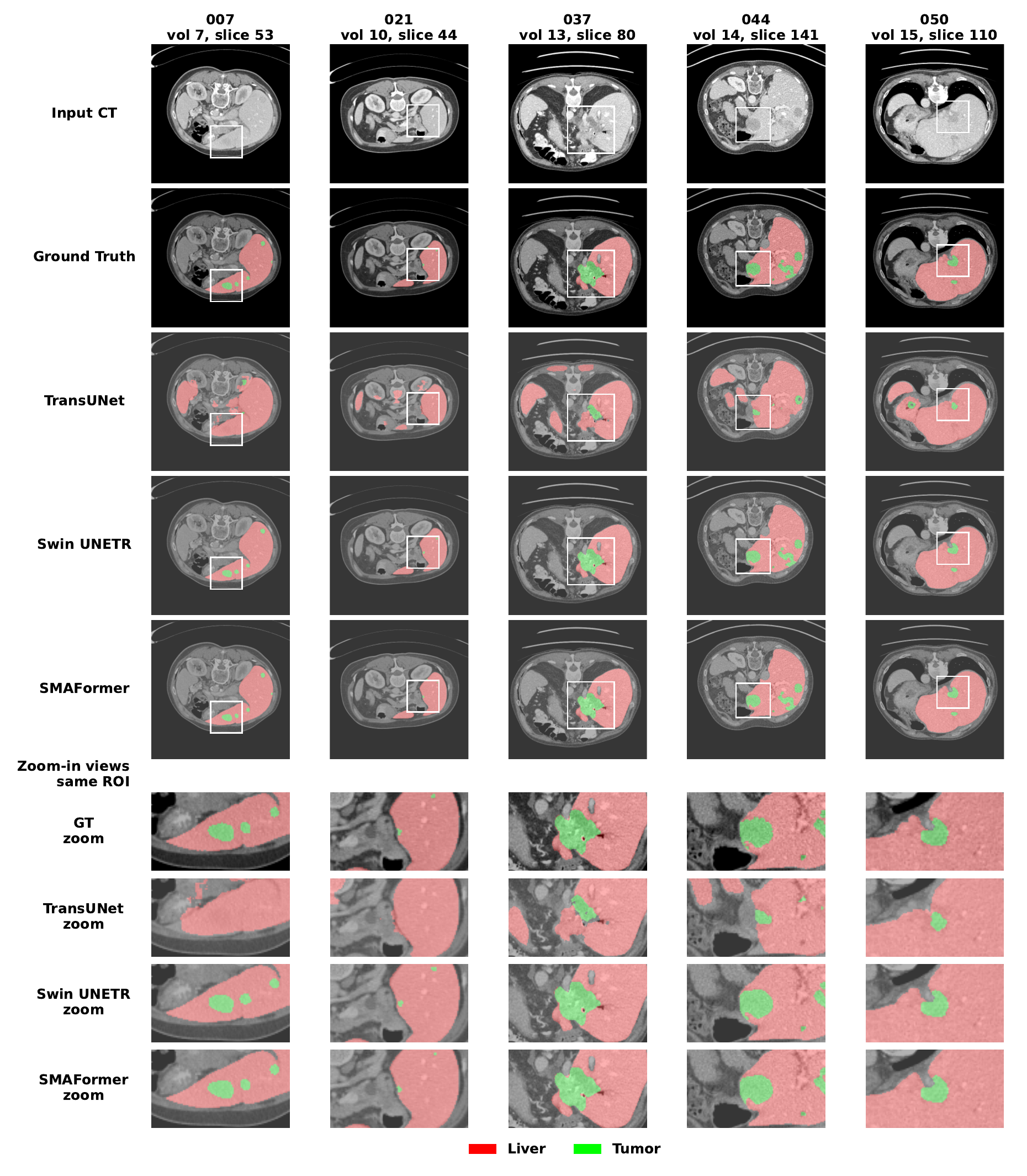}
\caption{Representative LiTS2017 liver tumor segmentation results. The examples emphasize small tumor regions where preserving local detail is critical.}
\label{fig3}
\end{figure}

Figure~\ref{fig:nature_smaformer_average} summarizes the tumor-dataset comparison from Table~\ref{tab:comp}. The point is to make the paired Dice and mIoU evidence visible rather than to repeat the table. SMAFormer reports average Dice values that are 1.51 points higher on ISICDM2019 and 0.83 points higher on LiTS2017; the corresponding mIoU differences are 1.07 and 1.13 points. The visual pattern is consistent across both overlap metrics, which supports the interpretation that the multi-attention and modulated fusion design improves tumor segmentation quality across both bladder and liver tumor settings under the reported protocol.

Figure~\ref{fig:nature_smaformer_synapse_organs} provides the organ-level evidence for the Synapse result in Table~\ref{tab:syna}. The point of the figure is that the performance profile is broad rather than concentrated in one organ. SMAFormer is highest or near-highest across large structures such as liver and spleen and remains competitive on smaller or more variable structures such as gallbladder and pancreas. The figure therefore links the average Dice gain to an organ-wise pattern rather than treating the average score as an isolated number.

\subsubsection{Liver Tumor Segmentation (Table~\ref{tab:comp})}

On LiTS2017, SMAFormer obtains the highest reported average scores in Table~\ref{tab:comp} (DSC: 94.11\%, mIoU: 91.94\%). Compared with TransUNet (DSC: 93.28\%, mIoU: 90.81\%), the result supports the value of combining synergistic multi-attention with modulated feature fusion for small and irregular tumors.

\subsubsection{Bladder Tumor Segmentation (Table~\ref{tab:comp})}

On ISICDM2019, SMAFormer achieves 96.07\% average DSC and 94.67\% mIoU. For the tumor class, it improves DSC by 2.18 percentage points over the second-best model. This improvement is consistent with the role of E-MLP in preserving local boundary context within the SMA Transformer block.

\subsubsection{Multi-Organ Segmentation (Table~\ref{tab:syna})}

On Synapse, SMAFormer achieves the strongest reported average DSC in Table~\ref{tab:syna} (86.08\%) and the highest DSC for five of eight organs. The organ-wise results indicate that the proposed design is most useful when local organ detail and global anatomical context must be balanced.

\subsection{Visualization of Segmentation Results}
\begin{figure}[ht]
\centering
\includegraphics[width=\columnwidth]{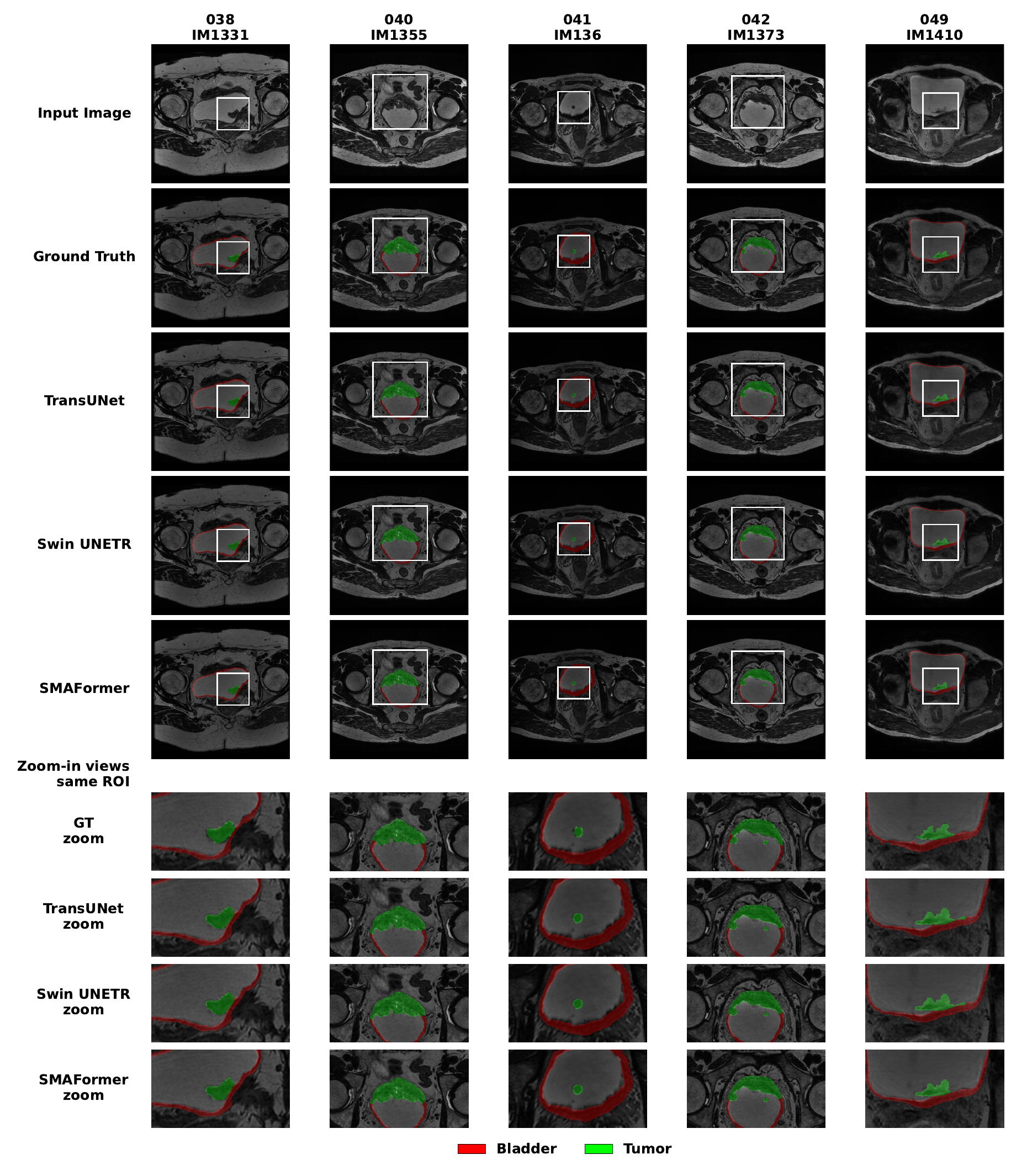}
\caption{Representative ISICDM2019 bladder tumor segmentation results. SMAFormer is designed to reduce the loss of irregular tumor boundary information during deep feature extraction.}
\label{fig4}
\end{figure}

\begin{figure}[ht]
\centering
\includegraphics[width=\columnwidth]{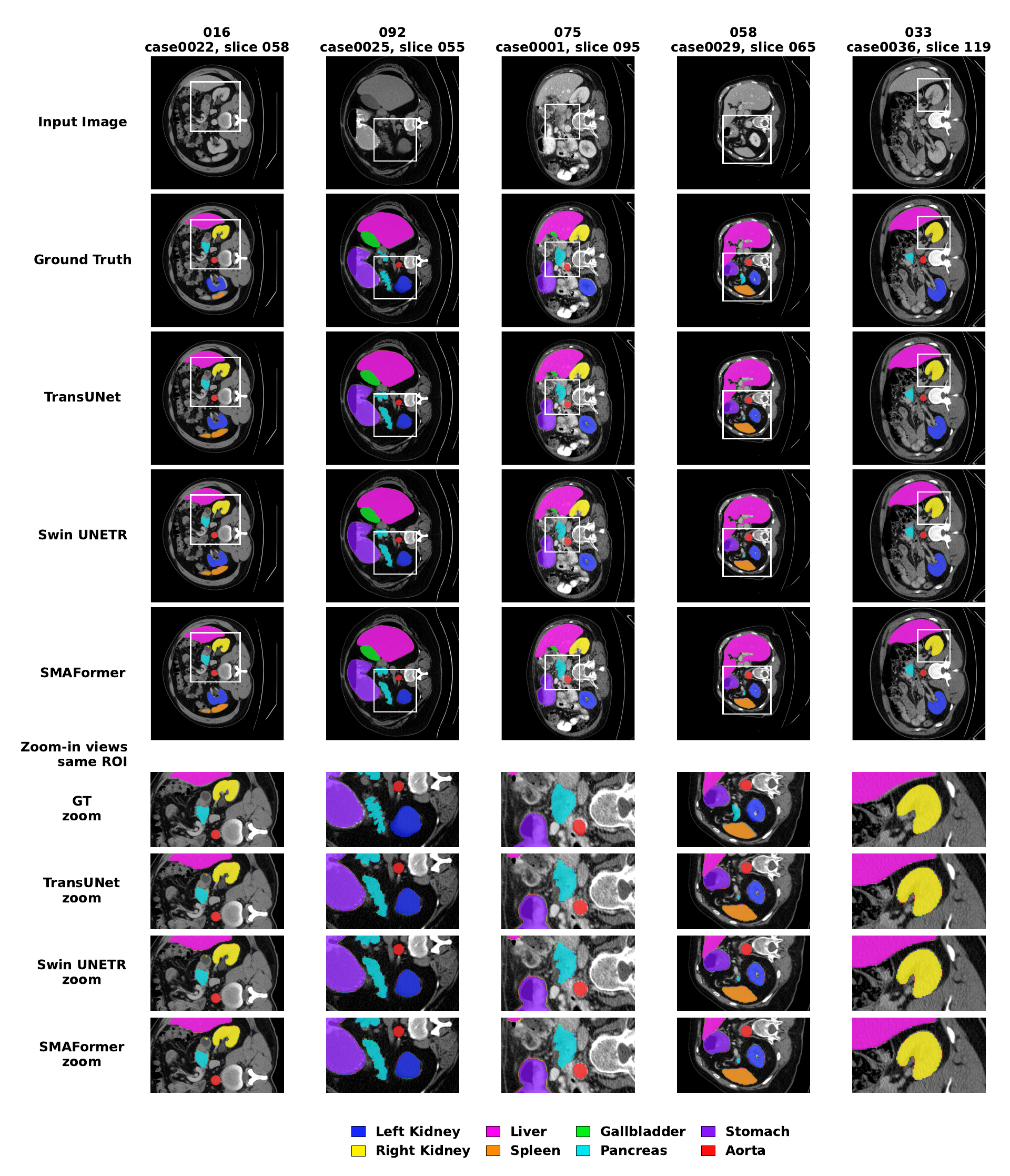}
\caption{Representative Synapse multi-organ segmentation results. The qualitative comparison illustrates the role of multi-scale attention in preserving small organs while maintaining global anatomical consistency.}
\label{fig5}
\end{figure}

Figure~\ref{fig3} shows representative LiTS2017 liver tumor cases. The panel inspects whether the tumor-class Dice in Table~\ref{tab:comp} corresponds to visible boundary behavior. In the shown slices, SMAFormer preserves compact tumor regions inside the liver instead of smoothing them into background or merging them with surrounding liver tissue. This behavior matches the quantitative evidence: LiTS2017 tumor Dice reaches 92.34\%, and the average Dice improves over the strongest listed baseline. The panel therefore supports the interpretation that E-MLP local context and attention fusion help retain small intra-organ evidence.

Figure~\ref{fig4} examines ISICDM2019 bladder tumor segmentation. The point is to test irregular tumor contours under a different anatomical site. The visual examples show more complete foreground regions and fewer boundary gaps around bladder tumor areas, consistent with the 93.56\% tumor Dice and 94.67\% mIoU in Table~\ref{tab:comp}. The mechanism-level reading is that pixel attention supplies fine localization, while the Feature Fusion Modulator reduces the loss of weak tumor-boundary evidence during channel-spatial transitions.

Figure~\ref{fig5} extends the qualitative check to multi-organ CT segmentation. The Synapse examples show coherent organ contours across adjacent abdominal structures, which aligns with the 86.08\% average Dice and the high liver, spleen, kidney, and stomach scores in Table~\ref{tab:syna}. The visual evidence indicates that the skip-connected U-shaped architecture still preserves global layout while SMA blocks refine local saliency.

\begin{figure}[!t]
    \centering
    \includegraphics[width=0.98\columnwidth]{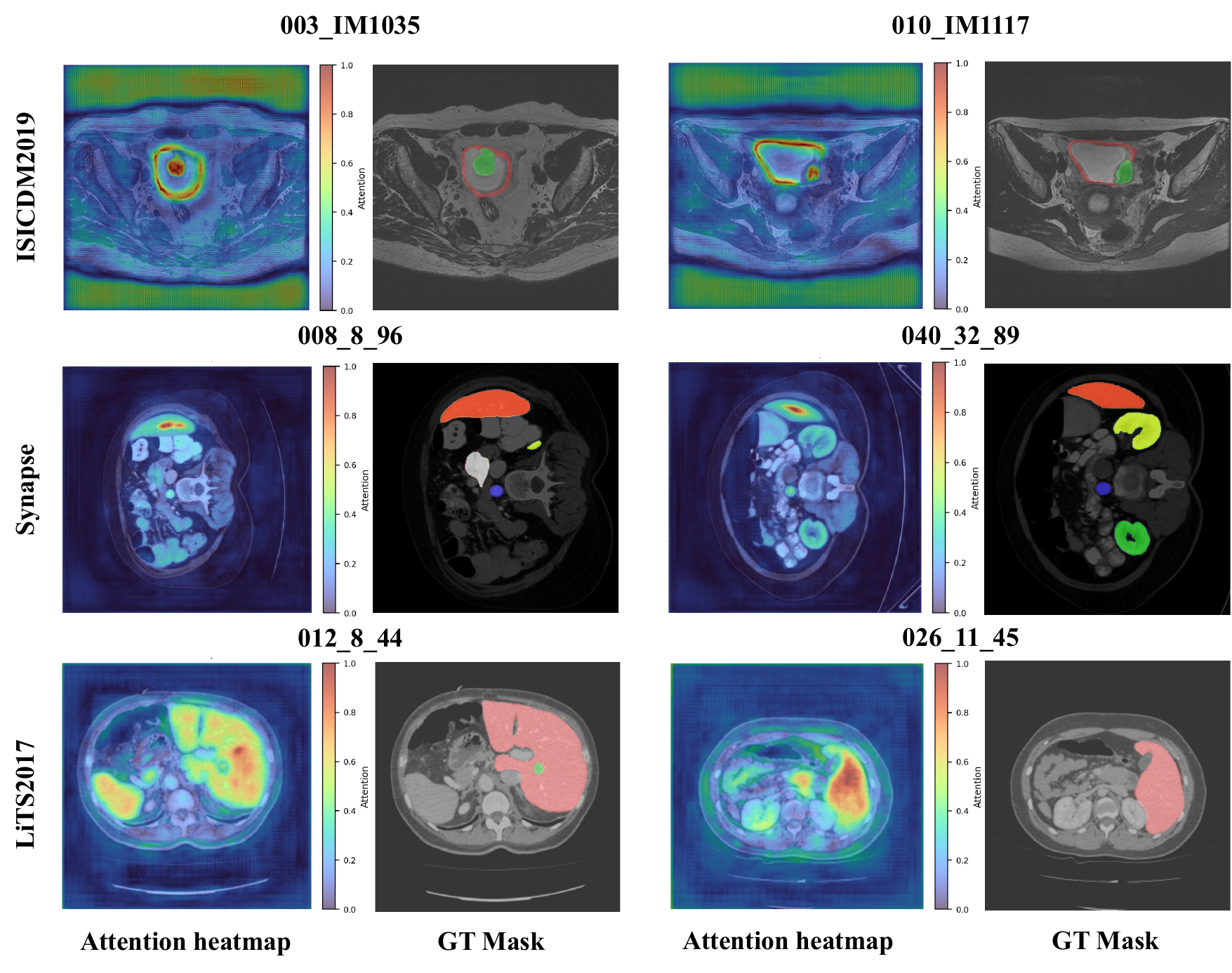}
    \caption{Attention heatmaps on representative ISICDM2019, Synapse, and LiTS2017 cases. High-response regions concentrate around tumor or organ sites, providing qualitative support for the multi-attention fusion behavior reported in the quantitative results.}
    \label{fig:attention_heatmaps}
\end{figure}

Figure~\ref{fig:attention_heatmaps} adds an attention-level diagnostic for SMAFormer. The point is to examine whether the model's visual evidence is concentrated near medically relevant structures. Across representative ISICDM2019, Synapse, and LiTS2017 examples, the heatmaps place high-response regions around tumor or organ sites rather than uniformly across the slice. This observation is consistent with the quantitative pattern in Tables~\ref{tab:comp}--\ref{tab:ablation}: the model improves small or difficult foreground regions when all attention and fusion components are retained. The heatmaps do not replace quantitative evaluation, but they provide visual evidence that the architecture changes where the model attends during segmentation.

In summary, the quantitative and qualitative results support the intended division of roles: the SMA Transformer block coordinates multiple attention views, the E-MLP preserves local evidence, and the feature fusion modulator reduces information loss during attention transitions.

\subsection{Ablation Study}

This subsection examines the impact of each component within SMAFormer through an ablation study on the ISICDM2019 and LiTS2017 datasets (experimental setup identical to Section~\ref{sec:4a}). Table~\ref{tab:ablation} summarizes the results.

\begin{table}[ht]
    \centering
    \caption{Ablation study of different modules in SMAFormer.}
    \resizebox{\columnwidth}{!}{%
    \begin{tabular}{c c c|c|c}
        \toprule
        SMA & E-MLP & Modulator &
        ISICDM2019 & 
        LiTS2017 \\
        & & &  Average DSC $\uparrow$ & Average DSC $\uparrow$ \\
        
        \midrule
    
            \checkmark & \ding{55} & \ding{55} & 82.28\% & 79.95\% \\
        
            \ding{55} & \checkmark & \ding{55} & 80.54\% & 75.67\% \\
        
            \ding{55} & \ding{55} & \checkmark & 78.41\% & 73.20\% \\

            \checkmark & \checkmark & \ding{55} & 89.53\% & 88.47\% \\

            \checkmark & \ding{55} & \checkmark & 86.31\% & 84.26\% \\

            \checkmark & \checkmark & \checkmark & 96.07\% & 94.61\% \\
        \bottomrule
    \end{tabular}
    }
    \label{tab:ablation}
\end{table}

\begin{figure}[t]
    \centering
    \includegraphics[width=\columnwidth]{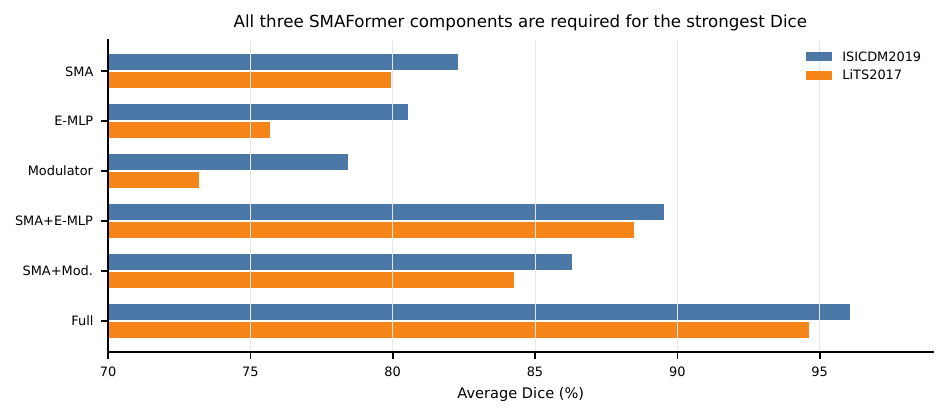}
    \caption{Component-level ablation from Table~\ref{tab:ablation}. The full design, combining SMA, E-MLP, and the modulator, produces the strongest average Dice on both datasets.}
\label{fig:nature_smaformer_ablation}
\end{figure}

Table~\ref{tab:ablation} studies whether the three proposed components act as independent add-ons or as a coordinated design. Under this component-removal protocol, the complete configuration gives the strongest tumor segmentation result. On ISICDM2019, single-component variants range from 78.41\% to 82.28\% average Dice, partial combinations reach 86.31\% or 89.53\%, and the full SMA + E-MLP + Modulator design reaches 96.07\%. On LiTS2017, the same ordering appears: the full model reaches 94.61\%, while the strongest partial variant reaches 88.47\%. The evidence supports a configuration-level interpretation: attention alone is not enough, local MLP context alone is not enough, and the modulator alone is not enough under the tested setting.

Figure~\ref{fig:nature_smaformer_ablation} visualizes the same ablation pattern and makes the component interaction easier to inspect. In both datasets, the full row is separated from every single-component and partial-combination row by a visible gap, especially on LiTS2017 where E-MLP alone reaches 75.67\% but the full model reaches 94.61\%. This pattern is consistent with the proposed division of roles: SMA supplies multi-view attention, E-MLP supplies local neighborhood processing, and the modulator preserves useful information during fusion.

\section{Conclusion}
We presented SMAFormer, a Transformer-based architecture for medical image segmentation tasks that require both fine local evidence and broader anatomical context. Its Synergistic Multi-Attention block integrates pixel, channel, and spatial attention to preserve local features while retaining global context. The multi-scale segmentation modulator further reduces information loss when attention representations are reshaped and fused across encoder-decoder scales.

Experiments on three public medical image segmentation datasets show that SMAFormer improves the reported benchmark results for liver tumor, bladder tumor, and multi-organ segmentation. These results support SMAFormer as a useful architecture for segmenting small organs and irregular tumors, while future work should further validate its behavior under broader clinical imaging protocols.

\FloatBarrier

\section{Acknowledgment}
This work was supported in part by the University of Macau under Grant MYRG2022-00190-FST, and in part by the Science and Technology Development Fund, Macau SAR, under Grant 0141/2023/RIA2 and 0193/2023/RIA3, in part by the Natural Science Foundation of Guangdong Province (No. 2023A1515010673), and in part by the Shenzhen Technology Innovation Commission (No. JSGG20220831110400001), and in part by Shenzhen Medical Research Fund (No. D2404001), and in part by Shenzhen Engineering Laboratory for Diagnosis \& Treatment Key Technologies of Interventional Surgical Robots (XMHT20220104009), and The Key Laboratory of Biomedical Imaging Science and System, CAS, for the Research platform support.

\FloatBarrier

\clearpage
\bibliographystyle{IEEEtran}
\bibliography{ref}

@article{1,
	title        = {Cancer immunoediting: from immunosurveillance to tumor escape},
	author       = {Dunn, Gavin P and Bruce, Allen T and Ikeda, Hiroaki and Old, Lloyd J and Schreiber, Robert D},
	year         = 2002,
	journal      = {Nature immunology},
	volume       = 3,
	number       = 11,
	pages        = {991--998}
}

@article{2,
	title        = {Medical image analysis},
	author       = {Ritter, Felix and Boskamp, Tobias and Homeyer, Andr{\'e} and Laue, Hendrik and Schwier, Michael and Link, Florian and Peitgen, H-O},
	year         = 2011,
	journal      = {IEEE pulse},
	volume       = 2,
	number       = 6,
	pages        = {60--70}
}

@article{3,
	title        = {A combined PET/CT scanner for clinical oncology},
	author       = {Beyer, Thomas and Townsend, David W and Brun, Tony and Kinahan, Paul E and Charron, Martin and Roddy, Raymond and Jerin, Jeff and Young, John and Byars, Larry and Nutt, Ronald},
	year         = 2000,
	journal      = {Journal of nuclear medicine},
	volume       = 41,
	number       = 8,
	pages        = {1369--1379}
}

@inproceedings{4unet,
	title        = {U-net: Convolutional networks for biomedical image segmentation},
	author       = {Ronneberger, Olaf and Fischer, Philipp and Brox, Thomas},
	year         = 2015,
	booktitle    = {MICCAI},
	pages        = {234--241}
}

@article{6resunet,
	title        = {ResUNet-a: A deep learning framework for semantic segmentation of remotely sensed data},
	author       = {Diakogiannis, Foivos I and Waldner, Fran{\c{c}}ois and Caccetta, Peter and Wu, Chen},
	year         = 2020,
	journal      = {ISPRS Journal of Photogrammetry and Remote Sensing},
	volume       = 162,
	pages        = {94--114}
}

@inproceedings{7resnet,
	title        = {Deep Residual Learning for Image Recognition},
	author       = {Kaiming He and Xiangyu Zhang and Shaoqing Ren and Jian Sun},
	year         = 2016,
	booktitle    = {CVPR},
	pages        = {770--778},
}

@inproceedings{12swin2021,
	title        = {Swin Transformer: Hierarchical Vision Transformer using Shifted Windows},
	author       = {Ze Liu and Yutong Lin and Yue Cao and Han Hu and Yixuan Wei and Zheng Zhang and Stephen Lin and Baining Guo},
	year         = 2021,
	booktitle    = {ICCV},
	pages        = {9992--10002},
}

@article{resT2022v2,
	title        = {Rest v2: simpler, faster and stronger},
	author       = {Zhang, Qinglong and Yang, Yu-Bin},
	year         = 2022,
	journal      = {NeurIPS},
	volume       = 35,
	pages        = {36440--36452}
}

@article{15chen2021transunet,
	title        = {Transunet: Transformers make strong encoders for medical image segmentation},
	author       = {Chen, Jieneng and Lu, Yongyi and Yu, Qihang and Luo, Xiangde and Adeli, Ehsan and Wang, Yan and Lu, Le and Yuille, Alan L and Zhou, Yuyin},
	year         = 2021,
	journal      = {ArXiv}
}

@inproceedings{21swinUnet2022,
	title        = {Swin-unet: Unet-like pure transformer for medical image segmentation},
	author       = {Cao, Hu and Wang, Yueyue and Chen, Joy and Jiang, Dongsheng and Zhang, Xiaopeng and Tian, Qi and Wang, Manning},
	year         = 2022,
	booktitle    = {ECCV},
	pages        = {205--218}
}

@inproceedings{34vit,
	title        = {An Image is Worth 16x16 Words: Transformers for Image Recognition at Scale},
	author       = {Alexey Dosovitskiy and Lucas Beyer and Alexander Kolesnikov and Dirk Weissenborn and Xiaohua Zhai and Thomas Unterthiner and Mostafa Dehghani and Matthias Minderer and Georg Heigold and Sylvain Gelly and Jakob Uszkoreit and Neil Houlsby},
	year         = 2021,
	booktitle    = {ICLR}
}

@article{44lits2017,
  title={The liver tumor segmentation benchmark (lits)},
  author={Bilic, Patrick and Christ, Patrick and Li, Hongwei Bran and Vorontsov, Eugene and Ben-Cohen, Avi and Kaissis, Georgios and Szeskin, Adi and Jacobs, Colin and Mamani, Gabriel Efrain Humpire and Chartrand, Gabriel and others},
  journal={Medical Image Analysis},
  volume={84},
  pages={102680},
  year={2023}
}

@inproceedings{45synapse,
	title        = {Miccai multi-atlas labeling beyond the cranial vault--workshop and challenge},
	author       = {Landman, Bennett and Xu, Zhoubing and Igelsias, J and Styner, Martin and Langerak, T and Klein, Arno},
	year         = 2015,
	booktitle    = {MICCAI},
	volume       = 5,
	pages        = 12
}

@proceedings{46ISICDM2019,
	title        = {Proceedings of the Third International                        Symposium on Image Computing and Digital Medicine, {ISICDM} 2019, Xi'an, China, August 24-26, 2019},
	year         = 2019,
	isbn         = {978-1-4503-7262-6}
}

@article{47nnformer,
	title        = {nnFormer: volumetric medical image segmentation via a 3D transformer},
	author       = {Zhou, Hong-Yu and Guo, Jiansen and Zhang, Yinghao and Han, Xiaoguang and Yu, Lequan and Wang, Liansheng and Yu, Yizhou},
	year         = 2023,
	journal      = {IEEE transactions on image processing}
}

@inproceedings{48unetr,
	title        = {Unetr: Transformers for 3d medical image segmentation},
	author       = {Hatamizadeh, Ali and Tang, Yucheng and Nath, Vishwesh and Yang, Dong and Myronenko, Andriy and Landman, Bennett and Roth, Holger R and Xu, Daguang},
	year         = 2022,
	booktitle    = {WACV},
	pages        = {574--584}
}

@inproceedings{49swinUNETR,
	title        = {Self-supervised pre-training of swin transformers for 3d medical image analysis},
	author       = {Tang, Yucheng and Yang, Dong and Li, Wenqi and Roth, Holger R and Landman, Bennett and Xu, Daguang and Nath, Vishwesh and Hatamizadeh, Ali},
	year         = 2022,
	booktitle    = {CVPR},
	pages        = {20730--20740}
}

@inproceedings{52cotr,
	title        = {Cotr: Efficiently bridging cnn and transformer for 3d medical image segmentation},
	author       = {Xie, Yutong and Zhang, Jianpeng and Shen, Chunhua and Xia, Yong},
	year         = 2021,
	booktitle    = {MICCAI},
	pages        = {171--180}
}

@article{53resunet++,
	title        = {A comprehensive study on colorectal polyp segmentation with ResUNet++, conditional random field and test-time augmentation},
	author       = {Jha, Debesh and Smedsrud, Pia H and Johansen, Dag and de Lange, Thomas and Johansen, H{\aa}vard D and Halvorsen, P{\aa}l and Riegler, Michael A},
	year         = 2021,
	journal      = {IEEE journal of biomedical and health informatics},
	volume       = 25,
	number       = 6,
	pages        = {2029--2040}
}

@inproceedings{62BCEDiceLoss2016v,
  title={V-net: Fully convolutional neural networks for volumetric medical image segmentation},
  author={Milletari, Fausto and Navab, Nassir and Ahmadi, Seyed-Ahmad},
  booktitle={3DV},
  pages={565--571},
  year={2016}
}

@inproceedings{66wang2022uformer,
  title={Uformer: A general u-shaped transformer for image restoration},
  author={Wang, Zhendong and Cun, Xiaodong and Bao, Jianmin and Zhou, Wengang and Liu, Jianzhuang and Li, Houqiang},
  booktitle={CVPR},
  pages={17683--17693},
  year={2022}
}

@inproceedings{70MLP2021cvt,
  title={Cvt: Introducing convolutions to vision transformers},
  author={Wu, Haiping and Xiao, Bin and Codella, Noel and Liu, Mengchen and Dai, Xiyang and Yuan, Lu and Zhang, Lei},
  booktitle={ICCV},
  pages={22--31},
  year={2021}
}

@inproceedings{71eighboringpixels2021CVPR,
  title={Neighbor2neighbor: Self-supervised denoising from single noisy images},
  author={Huang, Tao and Li, Songjiang and Jia, Xu and Lu, Huchuan and Liu, Jianzhuang},
  booktitle={CVPR},
  pages={14781--14790},
  year={2021}
}

@article{72depth-wiseConv2021localvit,
  title={Localvit: Bringing locality to vision transformers},
  author={Li, Yawei and Zhang, Kai and Cao, Jiezhang and Timofte, Radu and Van Gool, Luc},
  journal={arXiv},
  year={2021}
}

@inproceedings{73depth-wiseConv2018mobilenetv2,
  title={Mobilenetv2: Inverted residuals and linear bottlenecks},
  author={Sandler, Mark and Howard, Andrew and Zhu, Menglong and Zhmoginov, Andrey and Chen, Liang-Chieh},
  booktitle={CVPR},
  pages={4510--4520},
  year={2018}
}

@inproceedings{74depth-wiseConv2021incorporating,
  title={Incorporating convolution designs into visual transformers},
  author={Yuan, Kun and Guo, Shaopeng and Liu, Ziwei and Zhou, Aojun and Yu, Fengwei and Wu, Wei},
  booktitle={ICCV},
  pages={579--588},
  year={2021}
}

@article{75GELU2016gaussian,
  title={Gaussian error linear units (gelus)},
  author={Hendrycks, Dan and Gimpel, Kevin},
  journal={arXiv},
  year={2016}
}

@inproceedings{77PretrainViT2020image,
  title={An image is worth 16x16 words: Transformers for image recognition at scale},
  author={Dosovitskiy, Alexey and Beyer, Lucas and Kolesnikov, Alexander and Weissenborn, Dirk and Zhai, Xiaohua and Unterthiner, Thomas and Dehghani, Mostafa and Minderer, Matthias and Heigold, Georg and Gelly, Sylvain and others},
  booktitle={ICLR},
  year={2020}
}

@inproceedings{chen1,
    author = {Chen, Xuhang and Cun, Xiaodong and Pun, Chi-Man and Wang, Shuqiang},
    booktitle = {ICASSP},
    pages = {1-5},
    title = {Shadocnet: Learning Spatial-Aware Tokens in Transformer for Document Shadow Removal},
    year = {2023},
}

@inproceedings{chen2,
    author = {Li, Zinuo and Chen, Xuhang and Pun, Chi-Man and Cun, Xiaodong},
    booktitle = {ICCV},
    pages = {12449-12458},
    title = {High-Resolution Document Shadow Removal via A Large-Scale Real-World Dataset and A Frequency-Aware Shadow Erasing Net},
    year = {2023}
}

@inproceedings{chen3, 
    title={Devignet: High-Resolution Vignetting Removal via a Dual Aggregated Fusion Transformer with Adaptive Channel Expansion}, 
    booktitle={AAAI}, 
    author={Luo, Shenghong and Chen, Xuhang and Chen, Weiwen and Li, Zinuo and Wang, Shuqiang and Pun, Chi-Man}, 
    year={2024},
}

@inproceedings{chen4,
    author = {Li, Zinuo and Chen, Xuhang and Wang, Shuqiang and Pun, Chi-Man},
    booktitle = {IJCAI},
    pages = {1160-1168},
    title = {A Large-Scale Film Style Dataset for Learning Multi-frequency Driven Film Enhancement},
    year = {2023}
}

@article{chen5,
    title = {Generative AI for brain image computing and brain network computing: a review},
    author={Gong, Changwei and Jing, Changhong and Chen, Xuhang and Pun, Chi Man and Huang, Guoli and Saha, Ashirbani and Nieuwoudt, Martin and Li, Han-Xiong and Hu, Yong and Wang, Shuqiang},
    journal = {Frontiers in Neuroscience},
    volume = {17},
    pages = {1203104},
    year = {2023}
}

@inproceedings{chen6,
    author = {Chen, Xuhang and Lei, Baiying and Pun, Chi-Man and Wang, Shuqiang},
    booktitle = {PRCV},
    pages = {16-26},
    title = {Brain Diffuser: An End-to-End Brain Image to Brain Network Pipeline},
    year = {2023}
}

@inproceedings{chen7,
    title={Dual-Hybrid Attention Network for Specular Highlight Removal},
    author={Xiaojiao Guo and Xuhang Chen and Shenghong Luo and Shuqiang Wang and Chi-Man Pun},
    booktitle={ACM MM},
    year={2024}
}

@article{chen8,
    title = {WavEnhancer: Unifying Wavelet and Transformer for Image Enhancement},
    journal = {Journal of Computer Science and Technology},
    volume = {39},
    number = {2},
    pages = {336-345},
    year = {2024},
    author = {Zinuo Li and Xuhang Chen and Shuna Guo and Shuqiang Wang and Chi-Man Pun}
}

@article{chen9,
  title={Medprompt: Cross-modal prompting for multi-task medical image translation},
  author={Chen, Xuhang and Pun, Chi-Man and Wang, Shuqiang},
  journal={arXiv},
  year={2023}
}

@article{chen10,
    title={MFDNet: Multi-Frequency Deflare Network for efficient nighttime flare removal},
    author={Jiang, Yiguo and Chen, Xuhang and Pun, Chi-Man and Wang, Shuqiang and Feng, Wei},
    journal={The Visual Computer},
    pages={1--14},
    year={2024}
}

@inproceedings{chen11,
    title={Generative ai enables eeg data augmentation for alzheimer’s disease detection via diffusion model},
    author={Zhou, Tong and Chen, Xuhang and Shen, Yanyan and Nieuwoudt, Martin and Pun, Chi-Man and Wang, Shuqiang},
    booktitle={ISPCE-ASIA},
    pages={1--6},
    year={2023}
}

@article{chen12,
    title={Weakly supervised semantic segmentation via saliency perception with uncertainty-guided noise suppression},
    author={Liu, Xinyi and Huang, Guoheng and Yuan, Xiaochen and Zheng, Zewen and Zhong, Guo and Chen, Xuhang and Pun, Chi-Man},
    journal={The Visual Computer},
    pages={1--16},
    year={2024}
}

@article{chen13,
    title={Psanet: prototype-guided salient attention for few-shot segmentation},
    author={Li, Hao and Huang, Guoheng and Yuan, Xiaochen and Zheng, Zewen and Chen, Xuhang and Zhong, Guo and Pun, Chi-Man},
    journal={The Visual Computer},
    pages={1--15},
    year={2024}
}

@article{chen14,
  title={RM-UNet: UNet-like Mamba with rotational SSM module for medical image segmentation},
  author={Hao Tang and Guoheng Huang and Lianglun Cheng and Xiaochen Yuan and Qi Tao and Xuhang Chen and Guo Zhong and Xiaohui Yang},
  journal={Signal, Image and Video Processing},
  year={2024}
}

@inproceedings{py1,
    title={Image Augmentation with Controlled Diffusion for Weakly-Supervised Semantic Segmentation},
    author={Wu, Wangyu and Dai, Tianhong and Huang, Xiaowei and Ma, Fei and Xiao, Jimin},
    booktitle={ICASSP},
    pages={6175--6179},
    year={2024},
}

@article{py2,
    title={Top-K Pooling with Patch Contrastive Learning for Weakly-Supervised Semantic Segmentation},
    author={Wu, Wangyu and Dai, Tianhong and Huang, Xiaowei and Ma, Fei and Xiao, Jimin},
    journal={arXiv},
    year={2023}
}

@article{py3,
    title={APC: Adaptive Patch Contrast for Weakly Supervised Semantic Segmentation},
    author={Wu, Wangyu and Dai, Tianhong and Chen, Zhenhong and Huang, Xiaowei and Ma, Fei and Xiao, Jimin},
    journal={arXiv},
    year={2024}
}

@inproceedings{zhang2,
    title={SpA-Former: An Effective and lightweight Transformer for image shadow removal},
    author={Zhang, Xiaofeng and Zhao, Yudi and Gu, Chaochen and Lu, Changsheng and Zhu, Shanying},
    booktitle={IJCNN},
    pages={1--8},
    year={2023},
}

@article{zhang4,
  title={Enlighten-anything: When segment anything model meets low-light image enhancement},
  author={Zhao, Qihan and Zhang, Xiaofeng and Tang, Hao and Gu, Chaochen and Zhu, Shanying},
  journal={arXiv},
  year={2023}
}

@article{zhang9,
    title={AFFNet: attention mechanism network based on fusion feature for image cloud removal},
    author={Shen, Runhan and Zhang, Xiaofeng and Xiang, Yonggang},
    journal={International Journal of Pattern Recognition and Artificial Intelligence},
    volume={36},
    number={08},
    pages={2254014},
    year={2022}
}

@article{li1,
  title={Monocular robust 3d human localization by global and body-parts depth awareness},
  author={Li, Haolun and Pun, Chi-Man},
  journal={IEEE Transactions on Circuits and Systems for Video Technology},
  volume={32},
  number={11},
  pages={7692--7705},
  year={2022}
}

@inproceedings{li2,
  title={Cee-net: complementary end-to-end network for 3d human pose generation and estimation},
  author={Li, Haolun and Pun, Chi-Man},
  booktitle={AAAI},
  volume={37},
  number={1},
  pages={1305--1313},
  year={2023}
}

\end{document}